\documentclass[]{article}
\usepackage{proceed2e}
\usepackage{amssymb}
\usepackage{algorithm}
\usepackage{algpseudocode}
\usepackage{pgfplots}

\usepackage{times}

\title{An Efficient Algorithm for Bayesian Nearest Neighbours} 

\author{ {\bf Giuseppe Nuti \thanks{{ }{ }Primarily at UBS Securities LLC, 1285 Ave. of the Americas, New York, NY 10019}} \\
University College London\\
Dept. of Computer Science\\
Gower Street, London WC1E 6BT, U.K.\\
\texttt{ucacnut@ucl.ac.uk}\\
}
\providecommand{\keywords}[1]{\textbf{\textit{Keywords---}} #1}

\begin{document}

\maketitle

\begin{abstract}
K-Nearest Neighbours (k-NN) is a popular classification and regression algorithm, yet one of its main limitations is the difficulty in choosing the number of neighbours. We present a Bayesian algorithm to compute the  posterior probability distribution for $k$ given a target point within a data-set, efficiently and without the use of Markov Chain Monte Carlo (MCMC) methods or simulation\textemdash alongside an exact solution for distributions within the exponential family. The central idea is that data points around our target are generated by the same probability distribution, extending outwards over the appropriate, though unknown, number of neighbours. Once the data is projected onto a distance metric of choice, we can transform the choice of $k$ into a change-point detection problem, for which there is an efficient solution: we recursively compute the probability of the last change-point as we move towards our target, and thus \emph{de facto} compute the posterior probability distribution over $k$. Applying this approach to both a classification and a regression UCI data-sets, we compare favourably and, most importantly, by removing the need for simulation, we are able to compute the posterior probability of $k$ exactly and rapidly. As an example, the computational time for the Ripley data-set is a few milliseconds compared to a few hours when using a MCMC approach.
\end{abstract}
\section{INTRODUCTION \& RELATED WORK}

Various authors have explored the idea of Bayesian k-NN algorithms, e.g.\cite{2008arXiv0802.1357C, Guo:2010:BAN:1753297.1753298}, and originally \cite{10.2307/3088801}. The simplicity and elegance of k-NN lends itself, at least intuitively, to a Bayesian setting where the aim is to allow the number of neighbours to vary depending on the data (as highlighted in ~\cite{Ghosh20063113}.) Practically all of the work has relied on Markov Chain Monte-Carlo methods in some form or other; the use of simulation circumvents the need to model the full joint probability distribution of the data and the number of neighbours for any target. In an attempt to avoid the use of simulation, the authors in \cite{DBLP:journals/corr/abs-1305-1002} have approximated the likelihood function, albeit at the expense of accuracy and portability to regression problems. More recently, as an alternative approach, the idea of hubness is explored in \cite{Tomasev:2011:PAN:2063576.2063919}.\\

In a somewhat distinct branch of Bayesian statistics, numerous studies have focused on estimating change-point probabilities for data where the generating process is presumed to vary over time. Initial works were focused on partition analysis for the entire data-set, often using MCMC simulation: \cite{10.2307/2335381, 10.2307/2986119, doi:10.1093/biomet/82.4.711} (which, interestingly, is loosely connected with the computational complexity of estimating the posterior probability of $k$ using  approaches based on simulation.) Yet it was not until the authors in \cite{2007arXiv0710.3742P} presented an on-line version of Bayesian change-point estimation (with an $O(n)$ computational complexity,) that change-point problems become easily estimated.\\

As discussed in detail by \cite{manocha2007empirical}, a probabilistic view of $k$ in nearest neighbour algorithms outperforms standard cross-validation approaches, though practitioners often avoid the probabilistic approach due to its reliance on MCMC methods of estimation. By using the algorithm presented in \cite{2007arXiv0710.3742P} applied to the data ordered by distance to our target coordinates, we can compute the exact probability distribution for $k$ specific to our target point.

\section{EFFICIENT BAYESIAN NEAREST NEIGHBOUR}

The idea of calibrating the number of neighbours to the data is centered around the notion that, within the appropriate neighbourhood, data points are similar, or, in other words, they are generated by the same process. It is indeed our goal to determine how many $k$ neighbours represent such \emph{appropriate} neighbourhood.\\

\subsection{DATA GENERATING PROCESS}

If we order the data using a distance measure of choice with respect to a target point, we have transformed our assumption into the idea that the data-generating process is shared for the first $k$ points closets to the target. As such, moving from the most distant point towards our target, the underlying process generating the data can vary with a known probability and, when a change occurs, such process is drawn from a known prior distribution. We aim to determine the probability of such change-point having occurred at the various intervals between neighbours \textemdash once we have reached our target datum.\\

This formulation is equivalent to the change-point analysis in \cite{2007arXiv0710.3742P}: we can recursively keep track of the historical change-point probability until we reach the target point. The resulting probability of a change-point having occurred at $k$ points from our target is indeed the probability of $k$ being the correct number of neighbours.

\subsubsection{A Simple Example} 

As a simple two dimensional classification problem, imagine we have the data depicted in Figure \ref{fig:simpleData}, with the respective order, based on the Euclidean distance, shown below in Fig \ref{fig:simpleDetails}.

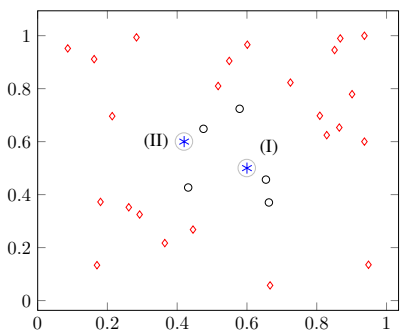
\begin{figure}[h]
    \centering
\begin{tikzpicture} [scale = 0.7]
	\begin{axis}[
	scatter/classes={%
		a={mark=o,draw=black},%
		b={mark=diamond,draw=red},%
		c={mark=asterisk,blue, mark size=3},
		d={mark=square*,orange, mark size=2}}]
	\addplot[scatter,only marks,%
		scatter src=explicit symbolic]%
	table[meta=label] {
x     y      label
0.6   0.5   c 
0.42   0.6   c
0.6544          0.4566      a
0.6010571622	0.965788727	b
0.7245673032	0.8228041473	b
0.9012435735	0.7790022197	b
0.9366343055	0.9995967383	b
0.808863355	0.6976428854	b
0.5173628945	0.8098629959	b
0.4453969078	0.2679131254	b
0.666278128	0.05730181278	b
0.9366323209	0.6000484448	b
0.5492957278	0.9045819851	b
0.4753643067	0.6483082918	a
0.1620806353	0.9113179007	b
0.2138134727	0.696091884	b
0.8511442833	0.9456421688	b
0.2924374165	0.3247738298	b
0.9482773183	0.1351774703	b
0.8674685266	0.9895811298	b
0.1702041479	0.1335612509	b
0.08642798335	0.951809558	b
0.662883892	0.3702780315	a
0.579029129	0.7239042902	a
0.828672222	0.6244064711	b
0.1802977616	0.3724057226	b
0.3645971478	0.2169889233	b
0.2834006093	0.9934884499	b
0.2612909393	0.3520839936	b
0.4315876843	0.4267978495	a
0.8646594311	0.6531626056	b
	};
	\coordinate (P1) at (513,443);
	\coordinate (P2) at (333,543);
	\node[draw,circle,label=above right:(I),color=lightgray] (CircleNode) at (P1) {};
	\node[draw,circle,label=left:(II),color=lightgray] (CircleNode) at (P2) {};
	\end{axis}
\end{tikzpicture}
\caption{Two-class data example; $k=5$ is most probably the correct chocie for target point (I) and any choice of $k\geq10$ will likely result in a misclassification, whilst the picture is less obvious for target point (II).}\label{fig:simpleData}
\end{figure}

 In this case, the appropriate number of neighbours for target point (I) is five. An alternative way to view the choice of $k$ is to order the data-points by their distance to the target point (Figure \ref{fig:simpleDetails}, below the x-axis.) We note the prior probability of each $k$ (before seeing any of the data) as the dotted line, which is just a geometric distribution with $p_\gamma\,{=}\,0.05$. To compute the posterior probability for $k$ (solid blue line in Fig. \ref{fig:simpleDetails},) we can start from the rightmost point and move towards our target: i.e. the probability that the data-generating process has changed (assuming a Beta prior probability for the data generation with parameters $\mathcal{B}(\alpha\,{=}\,10.,\beta\,{=}\,10.)$ and a probability of a change-point occurring in between any two neighbours of $p_\gamma\,{=}\,0.05$, i.e. our prior on the number of neighbours is $k=20$.) Conversely, we are not as convinced of the appropriate $k$ for target (II), as we can see form the posterior the distribution which is giving a rather mixed view. \\

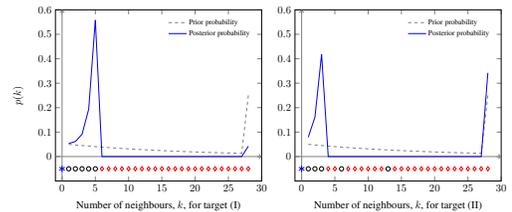
\begin{figure}[h]
    \centering
\begin{minipage}{.2\textwidth}
\begin{tikzpicture} [scale = 0.40]
\begin{axis}[ 
xlabel={Number of neighbours, $k$, for target (I)},
    ylabel={$p(k)$},
    xmin=-1, xmax=30,
    ymin=-.1, ymax=0.6,
    xtick={0,5,10,15,20,25,30},
    ytick={0,0.1,0.2,0.3,0.4,0.5,0.6,0.7,0.8,0.9,1.0},
    scatter/classes={%
		a={mark=o,draw=black},%
		b={mark=diamond,draw=red},%
		c={mark=asterisk,blue, mark size=3},
		d={mark=square,orange, mark size=3}},
    legend style={
        cells={anchor=west},
        draw=none, fill=none, 
        font=\scriptsize,
        legend pos= north east,}]
    
    \coordinate (h) at (axis cs:-1,0);
    \coordinate (i) at (axis cs:30,0);
    \coordinate (k) at (axis cs:0,-0.1);
    \coordinate (l) at (axis cs:0,0.6);
    \draw[->,gray,thick] (h) -- (i);
    \draw[->,gray,thick] (k) -- (l);
    


\addplot[
    color=gray,dashed]
    coordinates {
(1,0.05)
(2,0.0475)
(3,0.045125)
(4,0.04286875)
(5,0.0407253125)
(6,0.038689046875)
(7,0.03675459453125)
(8,0.0349168648046875)
(9,0.0331710215644531)
(10,0.0315124704862305)
(11,0.0299368469619189)
(12,0.028440004613823)
(13,0.0270180043831318)
(14,0.0256671041639753)
(15,0.0243837489557765)
(16,0.0231645615079877)
(17,0.0220063334325883)
(18,0.0209060167609589)
(19,0.0198607159229109)
(20,0.0188676801267654)
(21,0.0179242961204271)
(22,0.0170280813144058)
(23,0.0161766772486855)
(24,0.0153678433862512)
(25,0.0145994512169386)
(26,0.0138694786560917)
(27,0.0131760047232871)
(28,0.250344089742455)
    };

\addplot[
    color=blue]
    coordinates {
(1,0.053624)
(2,0.061537)
(3,0.091125)
(4,0.193472)
(5,0.557959)
(6,0)
(7,0)
(8,0)
(9,0)
(10,0)
(11,0)
(12,0)
(13,0)
(14,0)
(15,0)
(16,0)
(17,0)
(18,0)
(19,0)
(20,0)
(21,0)
(22,0)
(23,0)
(24,0)
(25,0)
(26,0)
(27,0)
(28,0.042283)

};

\addplot[scatter,only marks,%
		scatter src=explicit symbolic]%
	table[meta=label] {
x     y      label
0   -.05 c
1   -.05 a
2   -.05 a
3   -.05 a
4   -.05 a
5   -.05 a
6   -.05 b
7   -.05 b
8   -.05 b
9   -.05 b
10   -.05 b
11   -.05 b
12   -.05 b
13   -.05 b
14   -.05 b
15   -.05 b
16   -.05 b
17   -.05 b
18   -.05 b
19   -.05 b
20   -.05 b
21   -.05 b
22   -.05 b
23   -.05 b
24   -.05 b
25   -.05 b
26   -.05 b
27   -.05 b
28   -.05 b
};

\legend{Prior probability, Posterior probability}
\end{axis}
\end{tikzpicture}

\end{minipage}
\begin{minipage}{.2\textwidth}

\begin{tikzpicture} [scale = 0.40]
\begin{axis}[ 
xlabel={Number of neighbours, $k$, for target (II)},
    xmin=-1, xmax=30,
    ymin=-.1, ymax=0.6,
    xtick={0,5,10,15,20,25,30},
    ytick={0,0.1,0.2,0.3,0.4,0.5,0.6,0.7,0.8,0.9,1.0},
    scatter/classes={%
		a={mark=o,draw=black},%
		b={mark=diamond,draw=red},%
		c={mark=asterisk,blue, mark size=3},
		d={mark=square*,orange, mark size=2}},
    legend style={
        cells={anchor=west},
        draw=none, fill=none, 
        font=\scriptsize,
        legend pos= north east,}]
    
    \coordinate (h) at (axis cs:-1,0);
    \coordinate (i) at (axis cs:30,0);
    \coordinate (k) at (axis cs:0,-0.1);
    \coordinate (l) at (axis cs:0,0.6);
    \draw[->,gray,thick] (h) -- (i);
    \draw[->,gray,thick] (k) -- (l);
    


\addplot[
    color=gray,dashed]
    coordinates {
(1,0.05)
(2,0.0475)
(3,0.045125)
(4,0.04286875)
(5,0.0407253125)
(6,0.038689046875)
(7,0.03675459453125)
(8,0.0349168648046875)
(9,0.0331710215644531)
(10,0.0315124704862305)
(11,0.0299368469619189)
(12,0.028440004613823)
(13,0.0270180043831318)
(14,0.0256671041639753)
(15,0.0243837489557765)
(16,0.0231645615079877)
(17,0.0220063334325883)
(18,0.0209060167609589)
(19,0.0198607159229109)
(20,0.0188676801267654)
(21,0.0179242961204271)
(22,0.0170280813144058)
(23,0.0161766772486855)
(24,0.0153678433862512)
(25,0.0145994512169386)
(26,0.0138694786560917)
(27,0.0131760047232871)
(28,0.250344089742455)
    };

\addplot[
    color=blue]
    coordinates {
(1,0.07933)
(2,0.161097)
(3,0.417389)
(4,0)
(5,0)
(6,0)
(7,0)
(8,0)
(9,0)
(10,0)
(11,0)
(12,0)
(13,0)
(14,0)
(15,0)
(16,0)
(17,0)
(18,0)
(19,0)
(20,0)
(21,0)
(22,0)
(23,0)
(24,0)
(25,0)
(26,0)
(27,0)
(28,0.342184)
};

\addplot[scatter,only marks,%
		scatter src=explicit symbolic]%
	table[meta=label] {
x     y      label
0   -.05 c
1	-0.05	a
2	-0.05	a
3	-0.05	a
4	-0.05	b
5	-0.05	b
6	-0.05	a
7	-0.05	b
8	-0.05	b
9	-0.05	b
10	-0.05	b
11	-0.05	b
12	-0.05	b
13	-0.05	a
14	-0.05	b
15	-0.05	b
16	-0.05	b
17	-0.05	b
18	-0.05	b
19	-0.05	b
20	-0.05	b
21	-0.05	b
22	-0.05	b
23	-0.05	b
24	-0.05	b
25	-0.05	b
26	-0.05	b
27	-0.05	b
28	-0.05	b

};

\legend{Prior probability, Posterior probability}
\end{axis}
\end{tikzpicture}
\end{minipage}

\caption{Probability distribution for nearest neighbour count (with the data ordered by Euclidean distance below x-axis.) The prior probability is updated into the posterior probability by observing how the data will impact our choice of $k$ neighbours for the target (I) on the left and the target (II) on the right.}\label{fig:simpleDetails}
\end{figure}
 
 Note that the choice of Beta distribution is the standard conjugate prior for the parameters of a binary random variable. The choice of $p_\gamma\,{=}\,0.05$ is a key part of this approach. Specific to this example, we are implicitly assuming that the data will vary as we move away from our target point with a probability of 0.05. In other words, the hazard function is memoryless \textemdash \, with our prior for the expected number of neighbours set to $\frac{1}{p_\gamma}\,{=}\,20$.

\subsection{ALGORITHM}

 The first step is to represent the data into an ordered list driven by the distance from our target point\footnote{We can use any valid distance metric to produce such ordered list.}. If we define the target point as $x_\tau$, we order all of the available training data as $x_0,...,x_{\tau-1}$ (defined as $\vec{x}_{0:\tau-1}$) with $x_0$ being the most distant point from our target. We assume that the data $x_t$ is i.i.d. over a partition $\rho$ from some probability distribution $P(x_t|\eta_\rho)$, where $\eta_\rho$ represents the parameters of the data-generating distribution. Finally, we assume that for all of the partitions, $\eta_\rho$  is also i.i.d from a known prior distribution (where $\rho=1,...,n$ and $n\leq\tau$.) Now we are ready to use the algorithm presented in ~\cite{2007arXiv0710.3742P}, applied to the projected data\footnote{We note that the technique in \cite{2007arXiv0710.3742P} does introduce a slight approximation error, evident mainly for short run lengths. Alas, computing the exact posterior would increase the complexity of the algorithm to $O(n^2)$.}.\\
 
 Our objective is to compute the probability of each number of neighbours once we have reached our target point: $p(k_{\tau}{=}i|\vec{x}_{0,...,\tau-1}) \quad \forall i\,{=}\,0,...,\tau-1$ with $\tau-1$ total data points. Note that the subscript $\tau$ in $k_\tau$ indicates that we are representing the appropriate number of neighbours from the viewpoint of $x_\tau$, i.e. the target point. Starting from the point farthest away, and initializing the probability of a change-point having occurred before the initial point to 1.0, we set the initial conditions\footnote{Covered in more detail in the implementation notes.}:
 
  \begin{eqnarray}
    p(k_0=0) &=& 1.0\\
    \eta_0 &=& \eta_{prior}
 \end{eqnarray}
 
Firstly, we note that, as we observe a new datum, moving closer to our target, the number of neighbours $k_t$ within the same partition can either increase by one, with probability $1-p_\gamma$, or terminate in favour of a nascent partition.
 
  \begin{eqnarray}
    p(k_t|k_{t-1}) =  \left \{
  \begin{tabular}{ll}
  $p_\gamma$ & if  $k_t=0$  \\
  $1\,-\,p_\gamma$ & if $k_t=k_{t-1} + 1$\\
	0 & otherwise
  \end{tabular}
	\right.
 \end{eqnarray}

  A key advantage of the algorithm in \cite{2007arXiv0710.3742P} is that we can recursively compute the probability over the number of neighbours, $p(k_{t})$, by keeping track of the joint probability of each $k$ and the data: $p(k_{t-1}, x_0,...,x_{t-1})$, as we observe a new datum, $x_t$ alongside the predictive probability of $x_t$ for a given number of neighbours, $\pi_{t}=p(x_t|k_{t-1},\eta_{t-1})$.
  
   \begin{eqnarray}
    p(k_{t} = k_{t-1} + 1, x_0,...,x_t) = \qquad\qquad \qquad & \nonumber \\ 
    	\qquad\qquad\qquad p(k_{t-1}, x_0,...,x_{t-1})\, \pi_t \, p_\gamma \\
    p(k_{t} = 0, x_0,...,x_t) = \qquad\qquad\qquad\qquad \nonumber \\ 
    \qquad\qquad\qquad \sum_{k_{t-1}}p(k_{t-1}, x_0,...,x_{t-1})\, \pi_0 \, (1 - p_\gamma) 
 \end{eqnarray}
 
Finally, we define the notation $\eta_{\rho} \,{\twoheadleftarrow}\, x_t$ to indicate that we update the distribution parameters for $\eta_{\rho}$ with the datum $x_t$ using standard Bayesian updating rules\footnote{As a simple example, let's assume we are updating the probability of a Bernoulli distribution, e.g. a coin toss, with a prior of $\alpha=50$ for \emph{heads} and $\beta=50$ for \emph{tails} (where $\eta = \{ \alpha,\beta \}$ defined as a total of $100$ pseudo-observations and a prior probability $p(H)=0.5$.) If we then observe a new datum $x=\,$\emph{tails}, the $\eta \,{\twoheadleftarrow}\, x$ operation will update the parameters to  $\alpha'=50$ and $\beta'=51$, for a posterior predictive distribution of $p(H)=0.49505$.} (see \cite{Fink97acompendium} for conjugate prior updating within the exponential family.)\\

 %
\begin{algorithm}

\begin{algorithmic}
\State Initialize the data:
\State $\quad x_0,...,x_{\tau-1} \leftarrow$ ordered data for target point $\tau$
\State Initialize change-point variables:
\State $\quad p(k_0\,{=}\,0)\leftarrow 1.0$
\State $\quad \eta_0 \leftarrow \eta_{prior}$
\For {$t \leftarrow 0, \tau-1$}
\State Observe next variable $x_{t}$
\For {$i \leftarrow 0, t$}
	\State Compute predictive probability:
	\State $\quad \pi_i=p(x_t|k_{t-1}\,{=}\,i,\eta_i)$
	\State Compute growth probabilities:
	\State $\quad p(k_{t} \,{=}\, k_{t-1} + 1, \vec{x}_{0:t}) = p(k_{t-1}, \vec{x}_{0:t-1})\, \pi_t \, p_\gamma$
\EndFor	
\State Compute change-point probability:
\State $\quad p(k_{t} \,{=}\, 0, \vec{x}_{0:t}) =$
\State $\quad \quad \sum_{k_{t-1}}p(k_{t-1}, \vec{x}_{0:t-1})\, \pi_0 \, (1 - p_\gamma)$
\State Compute evidence:
\State $\quad p(\vec{x}_{0:t}) = \sum_{k_t} \; p(k_t,\vec{x}_{0:t})$
\For {$i \leftarrow 0, t$}
\State Compute probability of $k$:
\State $\quad p(k_t \,{=}\, i|\vec{x}_{0:t}) = \frac{p(k_t\,{=}\,i,\vec{x}_{0:t})}{p(\vec{x}_{0:t})}$
\State Update distributions:
\State $\quad \eta_{i} \,{\twoheadleftarrow}\, x_t$
\EndFor 
\EndFor \\
\Return $  p(k_\tau|\vec{x}_{0:\tau-1}) \quad \forall k_\tau \in \{0,..., \tau\}$
\end{algorithmic}
\label{algo:bknn}
\caption{Efficient Bayesian k-NN Algorithm}
\end{algorithm}

\subsubsection{Implementation Notes}
\renewcommand{\labelenumi}{(\alph{enumi})}
\begin{enumerate}
    \item The hazard function need not be constant; $p_\gamma\,{=}\,f(.)$ can, interestingly, depend on distance between points, or the current run-length (i.e. not memory-less,) etc.;
    \item Initializing the change-point probability: we do not have to set it to 1.0 before the first data-point. To speed up the analysis, we can start the algorithm $m$ points away from our target point (where $m$ can be set so that the prior probability of a change-point having occurred before $m$ falls below a preset threshold, and $m \,{\ll}\, n$.) In this case, and as an alternative, we can initialize the probability of a change-point before $m$ with the prior distribution for $p_{\gamma}$.
    \item The Bayesian update defined as '$\twoheadleftarrow$' can generally be computed efficiently for distributions in the exponential family. Other, possibly more complex, distribution may require a quadrature or simulation approach.
    \item For large data-sets, we resort to applying a $\log$ transform the joint probabilities in order to maintain numerical stability.
\end{enumerate}

\section{RESULTS}

In order to critically appraise this approach, we benchmark our analysis to the ubiquitous Ripley data-set for classification, and, as for a regression problem, to the Nuclear Power Plant output in \cite{kaya2012local}. The comparison is made against results obtained using the \emph{global} optimal number of neighbours (as a manual process.) In other words, we compare this approach against the best choice of $k$ when applied to all of the training data points. The key idea here is indeed that the optimal $k$ varies depending on the specific target point within the same data-set\footnote{Data and descriptions for both data-sets are
available the at UCI Machine Learning Repository (http://www.ics.uci.edu)}.

\begin{figure}[h]
    \centering
\begin{minipage}{.2\textwidth}
\begin{tikzpicture} [scale=0.45]
	\begin{axis}[
	scatter/classes={%
		a={mark=o,draw=red},%
		b={mark=diamond,draw=blue}}%
		]
	\addplot[scatter,only marks,%
		scatter src=explicit symbolic]%
	table[meta=label] {RipleyTraining.dat};
\end{axis}
\end{tikzpicture}
\end{minipage}
\begin{minipage}{.2\textwidth}
\begin{tikzpicture} [scale=0.45]
\begin{axis}[
	scatter/classes={%
		a={mark=o,draw=red},%
		b={mark=diamond,draw=blue}}%
		]
	\addplot[scatter,only marks,%
		scatter src=explicit symbolic]%
	table[meta=label] {RipleyTest.dat};
\end{axis}

\end{tikzpicture}
\end{minipage}
\caption{Ripley's traning data (left) and test data (right.)}\label{fig:RipleyData}
\end{figure}
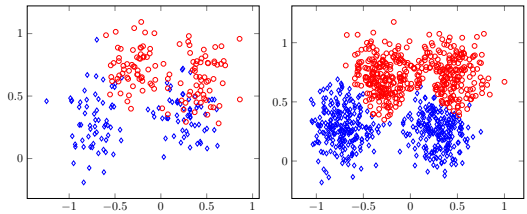

\begin{table}[h]
\begin{center}
\begin{tabular}{lcc}
\multicolumn{1}{c}{\bf Algorithm}  &\multicolumn{1}{c}{\bf Ripley}  &\multicolumn{1}{c}{\bf Power Plant}\\
\multicolumn{1}{c}{\bf }  &\multicolumn{1}{c}{\scriptsize (Misclassification)} &\multicolumn{1}{c}{\scriptsize (Avg. Abs. Error)}\\
\hline \\
k-NN (manual $k$ search)         &0.13 &3.6\\
Bayesian k-NN             &0.09 &2.9\\
\end{tabular}
\end{center}
\caption{Comparison of the global optimal $k$ neighbours versus the algorithm presented here: misclassification rate for Ripley data and average absolute error for Power Plant Output data.}
\label{TabRipleyResults}
\end{table}

In addition to the a prediction based on various $k$ values (weighted by their likelihood,) we now have a measure of certainty regarding our prediction. In Fig \ref{fig:RipleyPlot} we present the probability of classification computed using the training data.

\input{figRipleyPlot}

The MCMC-based Bayesian analysis in \cite{2008arXiv0802.1357C} achieved a misclassification rate of 0.087, which is, not surprisingly, similar to our result. We also note the similarity between our Figure \ref{fig:RipleyPlot} and the one presented in such study. Indeed, we are not proposing the idea of a Bayesian approach to estimating the number of neighbours, but an efficient method to do so. Using this algorithm, the time to compute the posterior distribution over $k$ for a test point within the Ripley data-set averaged three milliseconds per test point (for a standard PC,) compared to 50,000 paths used in the MCMC implementation in ~\cite{2008arXiv0802.1357C}. Notably, our approach results in the \emph{local} Bayesian analysis of $k$, i.e. specific to the data point being queried, as opposed to the global analysis for the MCMC approach.\\

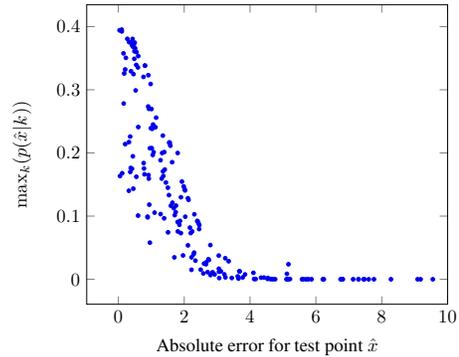
\begin{figure}[h]
    \centering
\begin{tikzpicture} [scale = 0.7]
\begin{axis}
[
legend style = { at = {(0.6,0.75)}},only marks,
xlabel = Absolute error for test point $\hat{x}$,xmax = 10,
ylabel = $\max_{k}(p(\hat{x}|k))$,
mark=o, mark size = 1]
\addplot table {MaxProbability.dat};
\end{axis}
\end{tikzpicture}
\caption{Maximum probability for 200 samples of realized test data across all possible $k$ neighbours (Power Plant Output data.)}
\label{fig:maxProbability}
\end{figure}

Finally, on the right column in Table \ref{TabRipleyResults} we show how, by switching the prior from a Beta distribution in the classification problem with the Normal distribution (as we only assumed the mean to be unknown,) we can obtain improved results for the Power Plant Output data. An interesting observation is that, if we plot the maximum posterior probability of the data w.r.t. the absolute error (in Fig. \ref{fig:maxProbability},) we can observe the ultimate limitation of k-NN algorithms. Data points with a large absolute error have clearly a very small probability of occurring, despite the fact that we are displaying the \emph{maximum} likelihood across all possible $k$'s; in other words, these are true outliers with respect to the distance measure that we have chosen\footnote{From an intuitive standpoint, we expect a plot of the maximum probability of the data across all values of $k$ to present outliers as having very low probability; this would indicate that the point is indeed dissimilar to \emph{any} grouping of neighbours for the chosen distance measure. In practice, a true outlier will have maximum probability for $k=0$, i.e. when it belongs to the uninformed prior as its the distribution with the largest variance, hence large absolute errors in Fig \ref{fig:maxProbability} converge onto a single line.}.

\section{CONCLUSIVE REMARKS}

We have presented an efficient algorithm to compute a Bayesian analysis over the number of neighbours in k-NN algorithms, applicable to classification and regression, which does not rely on MCMC simulation. This yields both superior predictions and a full probabilistic view of $k$. Yet the biggest challenge for k-NN algorithms is likely to be within the choice of the distance measure and differentiated input scaling (as highlighted in \cite{weinberger09distance}.) Certainly for multidimensional problems, the challenge lies in ordering the neighbours correctly with respect to their proximity to our target point, which in turn is driven by the coordinate transform we apply to compute the distance measure. An efficient Bayesian approach in understanding such scaling aspect may well be possible.



\bibliographystyle{unsrt} 
\bibliography{BK-NN}

\end{document}